\documentclass{easychair}

\usepackage{graphicx}
\usepackage{fancybox}
\usepackage{url}
\usepackage[utf8]{inputenc}
\usepackage{xspace}
\usepackage{amsmath}
\usepackage{amssymb}
\usepackage{booktabs}

\usepackage{setspace}
\usepackage{subfig}
\usepackage{paralist}
\usepackage{cite}
\usepackage{listings}
\usepackage{xcolor}

\newtheorem*{criterion}{Criterion}

\usepackage{hyperref}
\hypersetup{
  bookmarks=true,
  pdftitle={BliStr: The Blind Strategymaker}
  pdfauthor={Josef Urban}
  pdfkeywords={automated theorem proving, interactive theorem proving, machine learning, proof analysis, formal mathematics, mizar},
  colorlinks=true,
  linkcolor=black,
  citecolor=black,
}

\title{BliStr: The Blind Strategymaker}

\author{Josef Urban}
\institute{Radboud University   Nijmegen}

\titlerunning{BliStr: The Blind Strategymaker}

\authorrunning{Urban}

\begin{document}

\maketitle
\begin{abstract}
  BliStr is a system that automatically develops strategies for E
  prover on a large set of problems. The main idea is to interleave
  (i) iterated low-timelimit local search for new strategies on small
  sets of similar easy problems with (ii) higher-timelimit evaluation
  of the new strategies on all problems. The accummulated results of
  the global higher-timelimit runs are used to define and evolve the
  notion of ``similar easy problems'', and to control the selection of
  the next strategy to be improved. The technique was used to
  significantly strengthen the set of E strategies used by the
  MaLARea, PS-E, E-MaLeS, and E systems in the CASC@Turing 2012
  competition, particularly in the Mizar division. Similar improvement
  was obtained on the problems created from the Flyspeck
  corpus.
\end{abstract}
\section{Introduction and Motivation}
The E~\cite{Sch02-AICOMM} automated theorem prover (ATP) contains a number of
points where learning and tuning methods can be used to improve its
performance. Since 2006, the author has experimented with selecting
the best predefined E strategies for the Mizar/MPTP
problems~\cite{UrbanHV10,Urban06,Urb04-MPTP0}, and since 2011 the E-MaLeS~\cite{KuhlweinSU13} system has been
developed. This system uses state-of-the-art learning
methods to choose the best schedule of strategies for a problem.
An early evaluation of E-MaLeS in CASC 2011 has been counterintuitive:
E-MaLeS solved only one more FOF problem than E. Under reasonable
assumptions (imperfect knowledge, reasonably orthogonal strategies)
it is however easy to prove that for (super-)exponentially behaving
systems like E, even simple strategy scheduling should on average (and
with sufficiently high time limits) be better than running only one
strategy.  A plausible conclusion was that the set of E
strategies is not sufficiently diverse.

For the 2012 Mizar@Turing competition,\footnote{\url{http://www.tptp.org/CASC/J6/Design.html\#CompetitionDivisions}}
 1000 large-theory MPTP2078~\cite{abs-1108-3446}
problems (that would not be used in the competition) were released for
pre-competition training and tuning, together with their Mizar~\cite{mizar-in-a-nutshell} and
Vampire~\cite{Vampire} proofs. From the premises used in the Mizar proofs, Vampire
1.8 (tuned well for Mizar in 2010~\cite{UrbanHV10})\footnote{After writing the first version of this paper, Andrei Voronkov noted that he uses similar techniques for developing strategies for Vampire. A comparison of his system to BliStr would be interesting, but so far its description has not been published.}
could prove 691 of
these problems in 300s. A pre-1.6 version of E run
with its old auto-mode could solve only 518 of the problems.
In large-theory competitions like Mizar@Turing, where learning from
previous proofs is allowed, metasystems like MaLARea~\cite{US+08,malar14} can improve
the performance of the base ATP a lot. But the SInE premise-selection
heuristic~\cite{HoderV11} has been also tuned for
several years, and with the great difference of the base ATPs on
small version of the problems, the result of competition between
SInE/Vampire and MaLARea/E would be hard to predict. This provided a
direct incentive for constructing a method for automated improvement
of E strategies on a large set of related problems.

\section{Blind Strategymaking}
For the rest of this paper, the task (main criterion) is fixed to: 
\begin{criterion}[Max]
Develop a set of E
strategies that together solve as many of the 1000 small Mizar@Turing
pre-competition problems as possible.
\end{criterion}
\noindent A secondary criterion is that: 
\begin{criterion}[Gen]
The strategies should be reasonably general.
\end{criterion}
\noindent I.e., they should perform similarly also on
the unknown problems that would be later used in the competition. The third criterion is:
\begin{criterion}[Size]
The set of such strategies should not be too large.
\end{criterion}
This is to make sure that
strategy-selection systems like E-MaLeS stand a chance. This setting
is very concrete, however nothing particular is assumed about the
Mizar@Turing problems.

Even though the author has some knowledge of E (see, e.g.,~\cite{Urban06-ijait}), 
the strategy-improving methods were
intentionally developed in a data-driven way, i.e., assuming as little
knowledge about the meaning of E's strategies as possible.  The credo
of AI research is to automate out human intelligence, so rather than
manually developing deep theories about how the strategies work, which
ATP parameters are the right for tuning, how they influence each
other, etc., it was considered more interesting to push such a
``blind'' approach as far as possible, and try hard to automate the
discovery process based on data only. That is also why there is no
explanation of E strategies here (see the E manual), except the
following. 

A \textit{strategy} is assumed to be a collection of ATP parameters
with their (integer, boolean, enumerated) values. The parameters
influence the choice of inference rules, orderings, selection
heuristics, etc. Perhaps one unusual feature of E is that it provides
a language that allows the user to specify a linear combination of
clause-selection heuristics used during the given-clause loop. The
individual clause-selection heuristics further consist of a
(dependent) number of parameters, making the set of meaningful
strategy parameters very large (probably over 1000).  Capturing this
expressive power seemed tedious, and also looked like a hurdle to a
straightforward use of the established ParamILS~\cite{ParamILS-JAIR}
system which searches for good parameters by iterative local
search. Since ParamILS otherwise looks like the right tool for the
task, a data-driven (``blind'') approach was applied again to get a
smaller set (currently 20) of meaningful parameters: the existing
strategies that were (jointly) most useful on the training problems (see~\ref{choosing})
were used to extract a smaller set (a dozen) of clause-selection
heuristics. In some sense, an intelligent designer (Schulz) was trusted
to have already made reasonable choices in creating these smaller
building blocks, but we at least know that these are the building
blocks that provide the best performance so far, and reduce their
parameter search to their linear combinations. This can certainly be
made more ``blind'' later. The currently used set of parameters and
their allowed values\footnote{For the exact set of used parameters and values see the file e-params.txt in the BliStr distribution. The E interpretation of the parameters is in the file e\_wrapper1.rb .} limits the space of all expressible strategies to ca. $4.5*10^7$.

\subsection{Choosing a Starting Set of Strategies}
\label{choosing}
As mentioned above, the E auto-mode solves in 300s 518 of the 1000
problems.  One obvious choice of a set of starting strategies for
further improvement would be to take those that are used by the
auto-mode to solve the 518 problems.  The auto-mode is typically
constructed from an evaluation of about 280 pre-defined E strategies
on TPTP~\cite{Sutcliffe10}.  It has been observed several times that while the auto-mode
in general chooses good strategies on TPTP, it does not choose so well the
(still pre-defined) strategies for MPTP problems. In other words, even
though some MPTP problems are included in TPTP, the auto-mode should
not be trusted to know the best pre-defined strategies for MPTP. The
following method was used instead.

All the 280 pre-defined strategies were run on randomly chosen 200
problems from the 1000 with a low 5s timelimit, solving 117 problems
in total. A minimal set of strategies covering the 117 solutions was
computed (using MiniSat++), yielding the following six pre-defined 
strategies:\footnote{Their exact interpretation can be found in E's source code.}
\begin{small}
\begin{verbatim}
G-E--_008_B31_F1_PI_AE_S4_CS_SP_S2S  G-E--_008_K18_F1_PI_AE_CS_SP_S0Y 
G-E--_010_B02_F1_PI_AE_S4_CS_SP_S0Y  G-E--_024_B07_F1_PI_AE_Q4_CS_SP_S0Y
G-E--_045_B31_F1_PI_AE_S4_CS_SP_S0Y  G-E--_045_K18_F1_PI_AE_CS_OS_S0S    
\end{verbatim}
\end{small}
These six strategies were then run again on all the 1000 training
problems with 60s, proving 597 problems altogether. 
To get a fair (300s) comparison with the 300s runs of E and Vampire auto-mode,
only solutions obtained by each of these six strategies within 50s can
be considered. This yields 589 problems, i.e., a 13.7\% improvement
over the E auto-mode.  Thus, as conjectured, there are pre-defined E
strategies that can already do much better on MPTP problems than the E
auto-mode. However their difference from Vampire's performance (691
problems) is still large.

\subsection{Growing Better Strategies}

How can new strategies be found that would solve some of the unsolved
403 problems? 
The space of possible strategies is so large that a random
exploration seems unlikely to find good new 
strategies.\footnote{See Section~\ref{revolution} for an experiment in this direction.} 
The guiding idea is to again use a data-driven approach. Problems in a
given mathematical field often share a lot of structure and solution
methods. Mathematicians become better and better by solving the
problems, they become capable of doing larger and larger steps with
confidence, and as a result they can gradually attack problems that
were previously too hard for them. The reason for translating the
Mizar library for ATPs and having competitions like Mizar@Turing
is exactly to enable development and evaluation of systems that
try to emulate such self-improvement.

By this analogy, it is plausible to think that if the solvable problems
become much easier for an ATP system, the system will be able to solve
some more (harder, but related) problems. For this to work, a method
that can improve an ATP on a set of solvable problems is needed. While
this can still be hard (or even impossible), it is often much easier
task than to directly develop strategies for unsolved problems. The
reason is that an initial solution is known, and can be used as a
basis for algorithms that improve this solution using local search or other
non-random (e.g., evolutionary) methods. As already mentioned, the
established ParamILS system can be used for this.

\subsection{The ParamILS Setting and Algorithm}
Let $A$ be an algorithm whose parameters come from a
\textit{configuration space} (product of possible values) $\Theta$.  A
\textit{parameter configuration} is an element $\theta \in \Theta$,
and $A(\theta)$ denotes the algorithm $A$ with the parameter
configuration $\theta$.  Given a distribution (set) of problem
instances $D$, the \textit{algorithm configuration problem} is to find
the parameter configuration $\theta \in \Theta$ resulting in the best
performance of $A(\theta)$ on the distribution $D$.  ParamILS is an a
implementation of an \textit{iterated local search} (ILS) algorithm
for the algorithm configuration problem. In short, starting with an
initial configuration $\theta_0$, ParamILS loops between two steps: (i) perturbing the
configuration to escape from a local optimum, and (ii) iterative first
improvement of the perturbed configuration. The result of step (ii)
is accepted if it improves the previous best configuration. 

To fully determine how to use ParamILS in a particular case, $A$,
$\Theta$, $\theta_0$, $D$, and a performance metric need to be
instantiated. In our case, $A$ is E run with a low timelimit $t_{low}$, $\Theta$
is the set of the ca. $4.5*10^7$ E strategies, and as a performance
metric the number of given-clause loops done by E during solving the
problem was chosen. If E cannot solve a problem within the low
timelimit, a sufficiently high value ($10^6$) is used. CPU time
could be in some cases a better metric, however for very easy problems
it could be hard to measure the improvement factor with confidence.
It thus remains to instantiate $\theta_0$ and $D$.

\subsection{Guiding ParamILS}

It seems unlikely that there is one best E strategy for all
Mizar@Turing problems.  In principle this could be possible
particularly if the strategy language allowed to specify variant
behavior for different problem characterizations, however this is 
not yet the case.
Thus, it seems
counterproductive to use all the 597 solved training problems as the
set $D$ for ParamILS runs. If there is no best strategy, improved
performance of a strategy on one subset of all problems would be
offset by worse performance on another subset, the average value of
the performance metric would not improve, and ParamILS would not
develop such strategy further.

But this already suggests a data-driven way to guide ParamILS. If
there is no best strategy, then the set of all solvable problems is
partitioned into subsets on which the particular strategies perform
best. In more detail, this ``behavioral'' partitioning could be even
finer, and the vector of relative performances of all 
strategies on a problem could be used as a basis for various
clusterings of the problems. The current heuristic for
choosing successive $\theta_0$ and $D$ is as follows.  
\\
\\
\textbf{BliStr selection heuristic:} Let $\theta_i$ be a set of E
strategies, $P^j$ a set of problems, and $E_{\theta_i}(P^j)$ the
performance matrix obtained by running E with $\theta_i$ on $P^j$ with
a higher time limit $t_{high}$ (set to 10s). Let $c_{min} < c_{max}$
be the minimal and maximal eligible values of the performance metric
(given-clause count) (set to 500 and 30000). Let $E'_{\theta_i}(P^j)$
be $E_{\theta_i}(P^j)$ modified by using an $undef$ value for values
outside $[c_{min},c_{max}]$, and using an $undef$ value for all but
the best (lowest) value in each column.  Let $V$ (versatility) be the
minimal number (set to 8) of problems for which a strategy has to be
best so that it was eligible, and let $N$ be the maximum number of
eligible strategies (set to 20). Then the eligible strategies are the
first $N$ strategies $\theta_i$ for which their number of defined
values in $E'$ is largest and greater than $V$. These strategies are
ordered by the number of defined values in $E'$, i.e., the more the
better, and their corresponding sets of problems $D_i$ are formed by
those problems $P^j$, such that $E'_{\theta_i}(P^j)$ is defined.

Less formally, we prefer strategies that have many best-solvable
problems which can be solved within $[c_{min},c_{max}]$ given-clause
loops.  We ignore those whose versatility is less than $V$ (guarding
the \textit{Gen} criterion), and only consider the best $N$ (guarding
the \textit{Size} criterion). The maximum on the number of
given-clause loops guards against using unreasonably hard problems for
the ParamILS runs that are done in the lower time limit $t_{low}$
(typically 1s, to do as many ParamILS loops as possible). It is
possible that a newly developed strategy will have better performance
on a problem that needed many given-clause loops in the $t_{high}$
evaluation. However, sudden big improvements are unlikely, and using
very hard problems for guiding ParamILS would be useless. Too easy
problems on the other hand could direct the search to strategies that
do not improve the harder problems, which is our ultimate scheme for
getting to problems that are still unsolved. The complete BliStr loop
is then as follows. It iteratively co-evolves the set of strategies,
the set of solved problems, the matrix of best results, and the set of
eligible strategies and their problem sets.

\textbf{BliStr loop:} Whenever a new strategy $\theta$ is produced by
a ParamILS run, evaluate $\theta$ on all Mizar@Turing training
problems with the high time limit $t_{high}$, updating the performance
matrices $E$ and $E'$, and the ordered list of eligible strategies and
their corresponding problem sets. Run the next ParamILS iteration with
the updated best eligible strategy and its updated problem set, unless
the exact strategy and problem set was already run by ParamILS
before. If so, or if no new strategy is produced by the ParamILS run,
try the next best eligible strategy with its problem set. Stop when
there are no more eligible strategies, or when all eligible strategies
were already run before with their problem 
sets.\footnote{The stopping/selection criteria are now quite strict to see the limits of this approach. But it is easy to relax, e.g., by allowing further runs on smaller subsets, or letting survive/develop also  the ``not-best-enough'' strategies %
with high mean performance.}

This loop is implemented in about 500 lines of publicly available
Perl script.\footnote{\url{https://github.com/JUrban/BliStr}} It  implements the selection
heuristic, controls the ParamILS runs, and the higher-timelimit
evaluations. Content-based naming (SHA1) is used for the 
new strategies, so that many BliStr runs can be
merged as a basis for another run.

\section{Evaluation}

Table~\ref{bl1} summarizes two differently parametrized BliStr runs,
both started with the 6 pre-defined E strategies solving the 597
problems in 60s. Each BliStr run uses $t_{high}=10$ (which in
retrospect seems unnecessarily high). $BliStr_1^{400}$ uses
$t_{low}=1$ and a timelimit $T_{ParamILS}$ of 400s for each ParamILS
run. 37 iterations were done before the loop stopped. The 43 ($=6+37$)
strategies jointly cover 648 problems (when using $t_{high}$), the
best strategy solving 569 problems. Similarly for $BliStr_3^{2500}$,
which in much higher real time covered less problems, however produced
the strongest strategy.  Together with four other runs (some stopped
early due to an early bug, and some already start with some of the new
strategies), there were 113 ParamILS runs done in 30 hours of real
time on a 12-core Xeon 2.67GHz server, and covering 659 problems in
total (all with $t_{high}=10$).

\begin{table*}[htbp]
  \centering
  \caption{Two BliStr runs and a union of 6 runs done within 30 hours}
  \begin{tabular}{lrrrrrrr}
    \toprule
    description&$t_{low}$&$T_{ParamILS}$ &real time&user time&iterations&best strat.&solved\\\midrule
    $BliStr_1^{400}$ &1s&400s&593m&3230m&37&569&648\\
    $BliStr_3^{2500}$ &3s&2500s&1558m&3123m&23&576&643\\
    Union of 6 runs &&&1800m&&113&576&659\\\bottomrule    
  \end{tabular}
\label{bl1}
\end{table*}

22 strategies are (when using a simple greedy covering algorithm)
needed to solve the 659 problems, in general using $22 * 10s = 220s$, which
is less than the $6 * 60s =360s$ used by the 6 initial strategies to solve the
597 problems. These 22 strategies were later run also with a 60s time
limit, to have a comparison with the initial 6 strategies. Their joint
60s coverage is 670 problems. The (greedily) best 6 of them solve
together 653 problems, and the best of them solves 598 problems alone,
i.e. one more problem than the union of the initial strategies.
\\
\\
\textbf{Evaluation on Small Version of  the Mizar@Turing Problems:}
To see how general the strategies are, they were also evaluated on
small versions (i.e., using only axioms needed for their Mizar proof)
of the 400 Mizar@Turing competition problems, which were unknown
during the training on the 1000 pre-competition problems.  The
comparison in Table~\ref{bl2} includes the old E auto-mode, the 6
best pre-defined strategies, and Vampire 2.6 (used in the
competition). Each system was given 160s total CPU time (distributed
evenly between the strategies). The improvement over the old auto-mode is 25\%.

\begin{table*}[htbp]
  \centering
  \caption{Comparison on the (small) 400 competition problems using 160s}
  \begin{tabular}{lrrrr}
    \toprule
    System \  & \  Old auto-mode \  & \  6 old strats. \  & \  16 new strats. \  & \  Vampire 2.6\\\midrule
    Solved &205&233&253&273\\\bottomrule   
  \end{tabular}
\label{bl2}
\end{table*}

\textbf{CASC@Turing Competition Performance:}
An early version of a simple strategy scheduler and parallelizer
combining the best strategies also with (E's version of) SInE was used
by MaLARea in the Mizar@Turing competition. This strategy
scheduler\footnote{\url{https://github.com/JUrban/MPTP2/blob/master/MaLARea/bin/runepar.pl}}
(called Epar) runs 16 E strategies either serially or in
parallel.  In the competition MaLARea/Epar solved 257 of the 400
(large) Mizar@Turing problems in 16000 seconds, and Vampire/SInE 248 problems.\footnote{Vampire still won the competition: a bug in MaLARea caused 17 undelivered proofs.}
After the competition,
MaLARea was re-run on the 400 problems (on a different computer and 3 hours) both with Epar, solving 256 problems, and with
the old E's auto-mode, solving 214 problems. The
better E strategies were relevant for the 
competition.

The new strategies were also used by E-MaLeS and E1.6pre 
in the FOF@Turing competition run with 500 problems. E-MaLeS solved
401 of them, E1.6pre 378, and (old) E1.4pre 344. These improvements are
due to more factors (e.g., using SInE automatically in E1.6), however
the difference between E-MaLeS and E1.6pre became more visible in
comparison to the CASC 2011, likely also thanks to the
diverse strategies being now available.
\\
\\
\textbf{Evaluation on Flyspeck Problems:} Epar, E1.6pre and Vampire2.6
were also tested on the newly available Flyspeck
problems~\cite{holyhammer}. With 900s Vampire solves 39.7\% of all the
14195 problems, Epar solves 39.4\%, and their union solves
41.9\%. With 30s, on a random 10\% problem subselection, Epar solves
38.4\%, E1.6pre 32.6\%, and Vampire 30.5\% of the problems. This means
that on this completely different set of problems the newly developed
strategies solve 22\% more problems than the original version of E.

\section{Evolution vs. Revolution}
\label{revolution}
A.C. Clarke's Third Law states that any sufficiently advanced
technology is indistinguishable from magic. The evolutionary technique
described above is rational science, but without an explanation the
appearance of new strong strategies for an established ATP may look a
bit magical. In an early stage, the lack of explanation led one
colleague to suggest ``focusing on unsolved problems'' rather than
improving the performance on solved problems. While the science mostly
speaks against it (until there are improvable points, the search is
completely random and the search space is vast), a simple experiment was done later to
see how good this theory is. In this experiment, all the 403 unsolved
training problems were given to ParamILS, which was run for 7 hours,
starting with the default (bad) set of parameter values. 

The result of this long run was a strategy that solved 15 of the 403
problems.  The first success happened after about 1000 attempts.  The
likely explanation is that reasonable strategies perhaps are not so
rare in the constructed parameter space. Some parameters might have
relatively little importance once the more important parameters are
guessed reasonably well, thus effectively reducing the search for the
first successful data point. Even though this ``non-evolutionary''
approach is inferior to the evolutionary one, their combination might
bring further improvements.  This depends on how likely it is to randomly
hit a strategy that is good for a set of so far unsolved problems which are
relatively different from  all the problems solved so far.

\section{Conclusion and Future Work}
Running BliStr for 30 hours seems to be a good time investment for ATP
systems that are used to attack thousands of problems. It is also a
good investment in terms of the research time of ATP developers.  The
system can probably be made faster, and used online in metasystems
like MaLARea. The current selection heuristic could be modified in
various ways, as well as the stopping criterion.  The set of
parameters and their values could be extended, allowing broader
and more precise tuning. Extension to other ATPs should be straightforward.

\bibliographystyle{plain}
\bibliography{ate14}
\end{document}